\documentclass{article}
\usepackage{arxiv}

\usepackage[utf8]{inputenc} 
\usepackage[T1]{fontenc}    
\usepackage{hyperref}       
\usepackage{url}            
\usepackage{booktabs}       
\usepackage{amsfonts}       
\usepackage{nicefrac}       
\usepackage{microtype}      
\usepackage{lipsum}
\usepackage{tikz}
\usetikzlibrary{calc}

\usepackage{dsfont} 

\usepackage{amsmath}
\usepackage{amsthm}
\usepackage{amssymb}

\usepackage{changepage}
\usepackage{changes}
\usepackage{tabls}
\usepackage{afterpage} 
\usepackage{placeins}

\usepackage{subcaption,siunitx,booktabs}
\usepackage{adjustbox}
\usepackage{graphicx}
\usepackage{tikz}
\usepackage{lscape}
\usepackage{longtable} 
\usepackage{blindtext}
\usepackage{tabularx}
\usetikzlibrary{shapes,arrows,shapes.geometric}

\usepackage{algorithm}
\usepackage{algpseudocode} 
\algnewcommand{\LineComment}[1]{\State \(\triangleright\) #1}

\title{A Reinforcement Learning Method for Environments with Stochastic Variables: Post-Decision Proximal Policy Optimization with Dual Critic Networks\thanks{© 2025 IEEE. Personal use of this material is permitted. Permission from IEEE must be obtained for all other uses, in any current or future media, including reprinting/republishing this material for advertising or promotional purposes, creating new collective works, for resale or redistribution to servers or lists, or reuse of any copyrighted component of this work in other works.}}

\author{
  Leonardo Kanashiro Felizardo \\
  Division of Computer Science\\
  Instituto Tecnológico de Aeronáutica, 
  Brazil \\
  \texttt{felizardo@ita.br} \\
   \And
  Edoardo Fadda  \\
  Department of Mathematical Sciences\\
  Politecnico di Torino, 
  Italy \\
  \texttt{edoardo.fadda@polito.it} \\
  \AND
  Paolo Brandimarte \\
  Department of Mathematical Sciences\\
  Politecnico di Torino, 
  Italy \\
  \texttt{paolo.brandimarte@polito.it} \\
  \And
  Emilio Del-Moral-Hernandez \\
  Escola Politécnica\\
  Universidade de São Paulo, 
  Brazil \\
  \texttt{emilio.delmoral@usp.br} \\
  \AND
  Mariá Cristina Vasconcelos Nascimento \\
  Division of Computer Science\\
  Instituto Tecnológico de Aeronáutica, 
  São José dos Campos, Brazil \\
  \texttt{maria.nascimento@gp.ita.br} \\
}

\begin{document}
\maketitle

\begin{abstract}
This paper presents Post-Decision Proximal Policy Optimization (PDPPO), a novel variation of the leading deep reinforcement learning method, Proximal Policy Optimization (PPO). 
The PDPPO state transition process is divided into two steps: a deterministic step resulting in the post-decision state and a stochastic step leading to the next state. 
Our approach incorporates post-decision states and dual critics to reduce the problem's dimensionality and enhance the accuracy of value function estimation.
Lot-sizing is a mixed integer programming problem for which we exemplify such dynamics.
The objective of lot-sizing is to optimize production, delivery fulfillment, and inventory levels in uncertain demand and cost parameters.
This paper evaluates the performance of PDPPO across various environments and configurations. 
Notably, PDPPO with a dual critic architecture achieves nearly double the maximum reward of vanilla PPO in specific scenarios, requiring fewer episode iterations and demonstrating faster and more consistent learning across different initializations. 
On average, PDPPO outperforms PPO in environments with a stochastic component in the state transition.
These results support the benefits of using a post-decision state. 
Integrating this post-decision state in the value function approximation leads to more informed and efficient learning in high-dimensional and stochastic environments. 

\vspace{0.2cm}
\end{abstract}


\section{Introduction}
\label{sec:introduction}
Proximal Policy Optimization (PPO) \cite{Schulman2017} is a model-free RL method that has emerged as a leading deep Reinforcement Learning (RL) method, delivering state-of-the-art performance across a wide range of challenging tasks. 
Despite its success, PPO is highly reliant on the effectiveness of its exploratory policy search, which can be prone to risks such as insufficient exploration, particularly in cases of suboptimal initialization \cite{Zhang2022}. 
A core limitation lies in its exploration strategy, which relies heavily on Gaussian noise for action sampling. 
This approach often leads to suboptimal policy convergence, especially in environments where high-dimensional continuous action spaces and sparse rewards dominate.
Moreover, the stability and performance of PPO are highly dependent on the quality of sample-based estimates for policy gradients. 
High variance in these estimates, particularly in tasks with long time horizons, introduces instability during training.
This instability is further amplified in real-world applications, where data collection is expensive, and finite-sample estimation errors can have severe consequences \cite{Queeney2021}. 
These problems limit PPO's performance and restrict its scalability to more dynamic and uncertain environments, such as robotics, healthcare, and large-scale industrial simulations.

Despite these challenges, advancements have been proposed to address the limitations of PPO's exploration mechanisms.
For instance, methods proposed in \cite{Queeney2021, Zhang2022b} introduce mechanisms to enhance exploration and stability in PPO.
Zhang et al. \cite{Zhang2022b} address PPO's limited exploration capability by incorporating an Intrinsic Exploration Module (IEM) based on uncertainty estimation.
Similarly, Queeney et al. \cite{Queeney2021} propose the Uncertainty-Aware Trust Region Policy Optimization (UA-TRPO), which focuses on stabilizing policy updates under finite-sample estimation error.

This paper explores the limitations of PPO in two distinct environments. 
First, we analyze PPO's behavior in the Frozen Lake environment, a stochastic game scenario in which the agent must navigate a slippery grid world. 
This environment highlights PPO's reliance on effective exploration and showcases its struggles with stochastic transitions and sparse rewards, which are common challenges in many RL tasks. 
Also, Frozen Lake is an Atari game environment commonly employed in computer science research to train reinforcement learning agents.
Second, we investigate PPO in the Stochastic Discrete Lot-Sizing Problem (SDLSP) environment, which involves optimizing production and inventory decisions under random demand and uncertain cost parameters \cite{Felizardo2024}. 
This setting is a real-world industrial problem, where the combination of high-dimensional decision spaces and stochasticity tests the algorithm's stability and scalability.

To address some limitations of PPO in these types of environments, we propose the Post-Decision Proximal Policy Optimization (PDPPO), a novel variation of PPO that aims to enhance learning speed and results. 
This variation incorporates two distinct critic networks, one focused on learning the value of the state and another on learning the value of the post-decision state \cite{Powell2007}. 
The use of dual critic networks addresses the inherent limitations of traditional single-critic approaches, providing complementary perspectives that allow for better value function approximations. 
Moreover, by incorporating both state and post-decision state critics, PDPPO enhances the accuracy of the value function estimation, leading to better performance. More in detail, the post-decision state critic focuses on the post-decision state (i.e., the subsequent state after the decision's deterministic effects), effectively capturing the environment's deterministic dynamics. 
Instead, the state critic estimates the expected value of a given state by considering the current information. 

The enhanced value function estimation provided by the dual critic approach improves exploration capabilities and accelerates learning, as the agent can more effectively identify and prioritize high-reward actions. 
Consequently, PDPPO exhibits faster convergence and attains higher reward levels than methods relying on single-critic networks such as the vanilla PPO method.

The main contributions of this paper are:

\begin{itemize}
    \item We identify limitations in PPO and propose the PDPPO algorithm as a solution. The proposed method incorporates dual critic networks, leveraging state and post-decision state critics to enhance value function estimation and improve the efficiency of policy search.

    \item We tested the PDPPO algorithm in a modified Frozen Lake environment, separating the stochastic and deterministic effects into different steps. Testing in this environment provides more comparable performance results to evaluate the proposed method. The results suggest that PDPPO may perform better than standard PPO for this type of environment.
    
    \item We conduct experiments to evaluate the performance of PDPPO against vanilla PPO across various Stochastic Discrete Lot-Sizing environments. The results indicate that PDPPO, with its dual critic architecture, achieves superior learning speed and higher maximum reward levels.
    
\end{itemize}

\section{Related work}
\label{sec:related_work}

This section reviews the relevant literature on PPO, dual critic networks, and post-decision states approach, which serve as the foundation for the PDPPO algorithm.

\subsection{Proximal Policy Optimization}
\label{subsec:ppo}
PPO is an on-policy method that combines the advantages of trust region policy optimization \cite{Schulman2015} and clipped objective optimization. 
It balances exploration and exploitation by enforcing a constraint on the policy update to prevent overly large updates, which could lead to unstable learning. 
PPO has demonstrated success in a wide variety of complex tasks, including robotics \cite{Henderson2019}, inventory management \cite{Hezewijk2022}, telecommunication \cite{Hou2023}, etc.
Despite its success, PPO's reliance on a single critic network may limit its exploration capabilities and convergence speed.

Several studies have been conducted on different aspects of PPO in recent years. 
These include the impact of implementation choices \cite{Engstrom2020}, the ability to generalize \cite{Raileanu2021,Cobbe2020}, exploration \cite{Zhang2022}, and performance under multi-agent settings \cite{Yu2022}. 
Furthermore, some authors recognized that the value and policy learning tasks associated with actor-critic models such as PPO exhibit significant asymmetries \cite{Raileanu2021}.
Dual critic networks may address some general problems related to PPO and RL methods.

\subsection{Dual Critic Networks}

Several reinforcement learning algorithms have been explored using multiple critics to enhance learning performance. 
The Twin Delayed Deep Deterministic Policy Gradient algorithm \cite{Fujimoto2018} uses two critics in a continuous action space to address the overestimation bias in Q-value estimation. 
The Soft Actor-Critic (SAC) algorithm \cite{Haarnoja2018} also adopts a dual critic structure, combining it with an entropy-based exploration strategy to improve stability and exploration efficiency. 
More recently, other works employed different strategies with a dual critic approach \cite{Riashat2019,Cobbe2020,Koyejo2022} to improve learning in simulated environments. 
In real-world problems, there are examples of PPO using dual critic networks \cite{Wang2022}.
However, these methods have primarily been applied in cases with only one type of state variable, deterministic or stochastic.

The goal of dual critic networks is to use post-decision and pre-decision states.
Using post-decision states stems from \cite{Powell2007}, which suggests employing both states can improve value function estimation. 
The main idea behind dual critic networks is to leverage complementary information from both the current and subsequent states, thus offering a better understanding of the environment dynamics.

Additionally, incorporating post-decision state variables in dual critic methods enables a more precise depiction of the system's state at the decision point, reflecting the immediate outcomes of the decision before encountering any stochastic influences.
This approach leads to more informed and effective decisions that consider the environment dynamics' deterministic and stochastic components. 
Furthermore, post-decision value functions can help address the ``curse of dimensionality'' \cite{Bellman1957} problem when dealing with Q-factors \cite{Brandimarte2021}. 
Incorporating post-decision states and value functions can effectively reduce the problem's dimensionality by focusing on the relevant aspects of the system state at the moment of decision. This leads to faster and more efficient learning, especially in high-dimensional problems.
This approach has been employed in operations research and other types of problems to overcome the problems related to the curse of dimensionality as in \cite{Bertsekas2007,Senn2014,Hull2015,Zhang2023,Felizardo2024}.

To the best of our knowledge, this work is the first attempt to incorporate dual critic networks for post-decision states and standard states into the PPO method for sequential decision problems with uncertainty.  

\section{Preliminaries}
\label{sec:preliminaries}

A Markov Decision Process (MDP) is defined by  a tuple $(\mathcal{S}, \mathcal{A}, \mathcal{P}, \mathcal{R}, \gamma)$, where:

\begin{itemize}
    \item $\mathcal{S}$ is a finite set of states, where each state $s \in \mathcal{S}$ represents a specific configuration of the environment.
    \item $\mathcal{A}$ is a finite set of actions, where each action $a \in \mathcal{A}$ represents a decision that can be made by the agent in a given state.
    \item $\mathcal{P} : \mathcal{S} \times \mathcal{A} \times \mathcal{S} \rightarrow [0, 1]$ is the transition probability function, where $\mathcal{P}(s' | s, a)$ denotes the probability of transitioning from state $s$ to state $s'$ when taking action $a$.
    \item $\mathcal{R} : \mathcal{S} \times \mathcal{A} \times \mathcal{S} \rightarrow \mathbb{R}$ is the reward function, where $\mathcal{R}(s, a, s')$ represents the immediate reward obtained when transitioning from state $s$ to state $s'$ after performing action $a$.
    \item $\gamma \in [0, 1]$ is the discount factor, which determines the importance of future rewards relative to immediate rewards. A value of $\gamma$ close to $1$ means that future rewards are highly important, while a value close to $0$ indicates a preference for immediate rewards.
\end{itemize}

The value function $V^\pi(s)$ represents the expected cumulative reward starting from state $s$ and following policy $\pi$, while the action-value function $Q^\pi(s, a)$ extends this concept by considering the cumulative reward starting from state $s$, taking action $a$, and then following policy $\pi$. 
The policy $\pi(a|s)$ defines the agent's strategy by specifying the probability of taking action $a$ in state $s$. Together, these functions form the core of reinforcement learning, enabling the agent to effectively evaluate states, actions, and strategies.

The advantage function, $A^{\pi}(s, a)$, is the difference between the action-value function $Q^\pi(s, a)$ and the value function $V^\pi(s)$, thus providing a measure of the relative value of each action, helping to distinguish better actions than the average:
    \begin{equation}
        \ A^{\pi}(s, a) = Q^{\pi}(s, a) - V^{\pi}(s).
    \end{equation}

\section{Post-Decision Proximal Policy Optimization}\label{sec:proposed_method}





Let $V^{\pi,pre}(s)$ be the state value function estimated by the critic network and $V^{\pi,x}(s_t)$ be the post-decision state value function estimated by the second critic network. The advantages for each state in a trajectory are

\begin{equation}
\begin{aligned}[t]
A^{\pi,x}_t(s_t) = R^x_t - V^{\pi,x}(s_t) \qquad \text{and}\\
A^{\pi,pre}_t(s_t) = R_t - V^{\pi,pre}(s_t)
\end{aligned}
\end{equation}
where $R_t$ and $R^x_t$ are the discounted rewards and $r_t$ is the reward at time $t$, defined as:

\begin{equation}
R^x_t = \sum_{k=t}^{T} \gamma^{k-t} r^x_k, \qquad \text{and} \qquad R_t = \sum_{k=t}^{T} \gamma^{k-t} r_k.
\end{equation}


Note that $R_t$ is not computed as in the PPO since here it is associated with the deterministic reward right after the decision and before reaching the post-decision state. We then compute the advantage of a state as the maximum between the two advantages. In formula,

\begin{equation}
\label{eq:advantage_max}
A^{\pi}_t(s_t) = \max(A^{\pi,x}_t(s_t), A^{\pi,pre}_t(s_t))
\end{equation}

The purpose of choosing the maximum value between $A_t^{\pi, x}$ and $A_t^{\pi, p r e}$ is that the algorithm selects the advantage estimate that demonstrates the highest potential for improvement from a given state. 
This choice promotes selecting the most beneficial actions relative to the current policy, thereby accelerating the learning and adaptation process. Also, by optimizing for the maximum advantage, the algorithm may encourage the exploration of actions that could lead to higher rewards but are not frequently chosen under the current policy.

Finally, we update the policy ($\theta_\pi$) and critic networks ($\theta_{V}, \theta_{V^x}$) parameters using the following loss functions:

\begin{equation}
L(\theta_{V}) = MSE(V^{\pi,pre}(s_t),R(s_t)),
\label{eq:first_critic_pdppo_loss}
\end{equation}

\begin{equation}
L(\theta_{V^x}) = MSE(V^{\pi,x}(s^x_t),R(s^x_t)),
\label{eq:second_critic_pdppo_loss}
\end{equation}  

\begin{equation}
\begin{aligned}[t]
    L(\theta_\pi) = 
    & \min\big(\rho_t A^{\pi}_t, 
    \text{clip}(\rho_t, 1 - \epsilon, 1 + \epsilon) A^{\pi}_t\big) \\
    & + c_1 B[\pi_{\theta_\pi}(\cdot | s_t)] \\
    & + c_2 \big(L(\theta_{V}) + L(\theta_{V^x})\big),
\end{aligned}
\label{eq:actor_pdppo_loss}
\end{equation}

As in the PPO algorithm, the constants $c_1$ and $c_2$ are hyperparameters that weigh the contributions of the clipped surrogate objective and the entropy bonus to the actor's overall loss function. 
The entropy term $B[\pi_{\theta_\pi}(\cdot|s_t)]$ encourages exploration by adding stochasticity to the policy's decisions. 
The clipping parameter $\epsilon$ limits the extent to which the new policy can diverge from the old policy during updates, thereby stabilizing the learning process.

In this work, we leverage the concept of post-decision states and post-decision value functions \cite{Powell2007}. 
A post-decision state, denoted as $s^x$, is the system state right after a decision's deterministic effects but before the consequences of that decision fully unfold (exogenous variables). 
This state represents the system right after the decision point, excluding any subsequent stochasticity. 
In some environments, it can be seen as a parsimonious way of representing the couples $(s, a)$ \cite{Powell2007}. Mathematically, it is defined by:

\begin{equation}\label{eq:transition_det}
s^x = f(s, a)
\end{equation}
where $s$ is the system's current state, $a$ is the action chosen in $s$, and $f$ is the deterministic function that maps the current state and decision to the post-decision state. It is important to note that the proposed technique can be used only when the transition of the system from state $s$ to state $s'$ can be divided into two parts:
\begin{equation}
    s^x = f(s, a) \qquad \text{and} \qquad s' = s^x + \eta,
\end{equation}
where $\eta$ is some random factor. 

Leveraging post-decision state variables, we can define the \textit{post-decision value function} that estimates the expected discounted return starting from the post-decision state $s^x$ when following policy $\pi$. In formula

\begin{equation}\label{eq:pd_value}
 V^{\pi, x}(s^x)=\mathcal{R}(s^x)+\gamma \sum_{s^{\prime} \in \mathcal{S}} \mathcal{P}\left(s^{\prime} \mid s^x\right) V^{\pi,pre}\left(s^{\prime}\right).
\end{equation}

The post-decision value function is utilized to compute the post-decision advantage in \eqref{eq:advantage_max}, essential for updating the policy parameters. 
To calculate the loss employed in the parameter update of the post-decision value function, we sample $R(s^x_t)$ and calculate the mean squared error (MSE) between the return and the estimated values.

The described architecture of PDPPO leverages dual critics, each tailored to a specific state representation, needing different optimization strategies for each network to effectively minimize their respective loss functions. 
Consequently, different optimizers are employed for each critic to deal with their differences in terms of learning dynamics and convergence behaviors. 
This optimization approach ensures that each critic adapts optimally to its input data stream, enhancing the overall efficiency and effectiveness of the learning process.

\subsection{The algorithm}

The PDPPO algorithm initializes the policy and value networks with random weights and collects trajectories using the current policy. For each trajectory, it calculates the returns and advantages for each state.

The policy and value networks are optimized over $K$ epochs by sampling batches of transitions from the trajectories. 
For each batch, the algorithm computes log probabilities, state values, post-decision state values, advantages, importance sampling ratios ($\rho_t$), and the actor and critic losses. 
These losses update the networks using separate optimizers, unlike traditional PPO.

The mean PDPPO loss and its gradients with respect to the policy and value parameters are calculated, clipped by $\ell_2$ norm, and applied to update the networks. 
After $K$ epochs, the current policy parameters are synced with the old policy.

This process repeats for $N$ training epochs, and the trained policy parameters are returned. Algorithm \ref{algo:PDPPO} outlines the procedure.

\begin{figure}[ht]
    \centering
    \begin{tikzpicture}[node distance=2.5cm, auto] 

    \node [draw, rectangle, minimum width=9cm, minimum height=1cm] (env) {Environment}; 

    \node [draw, rectangle, below of=env, yshift=-0.5cm, minimum width=1.5cm, minimum height=1cm] (actor) {Actor};

    \node [draw, rectangle, below left of=actor, xshift=-1cm, yshift=0.5cm, minimum width=1.5cm, minimum height=1cm] (critic1) {Critic};

    \node [draw, rectangle, below right=of actor, xshift=-0.4cm, yshift=1.75cm, minimum width=1.5cm, minimum height=1cm, text width=1.5cm, align=center] (critic2) at (actor.south east) {Post-decision \\ critic};

    \draw [->] (env) -- node[pos=0.3, left]{$\textbf{s},\textbf{r}$} (critic1);
    \draw [->] (env) -- node[pos=0.3, right]{$\textbf{s}^x,\textbf{r}^x$} (critic2);
    \draw [->] ([xshift=-1.5ex]env.south) -- node[pos=0.3, above left, yshift=-0.8cm]{$\textbf{s},\textbf{r}$} ([xshift=-1.5ex]actor.north);
    \draw [->] ([xshift=+1.5ex]actor.north) -- node[pos=0.3, below right, yshift=0.2cm]{$\textbf{a}$} ([xshift=1.5ex]env.south);
    \draw [->] (critic1) -- node[pos=0.5, below]{$\textbf{V}$} (actor);
    \draw [->] (critic2) -- node[pos=0.5, below]{$\textbf{V}^x$} (actor);

    \draw[dashed, gray] 
  ($(critic1.north west) + (-0.9,1.7)$) 
  rectangle 
  ($(critic2.south east) + (0.8,-0.5)$);
\node [align=center, above right] at 
  ($(critic1.north west) + (-0.4,1.2)$) 
  {PDPPO agent};

    \end{tikzpicture}
    \caption{This figure depicts the architecture of the PDPPO agent used for training reinforcement learning models. The environment is evaluated by two critic networks: Critic and Post-decision Critic. The actor-network is updated by evaluating the value of the post-decision critic and critic. The agents use a batch sample of $\textbf{s}$, $\textbf{r}$, $\textbf{s}^x$, and $\textbf{r}^x$ to update both critic and post-decision critic.}
    \label{fig:PDPPO_architecture}
\end{figure}
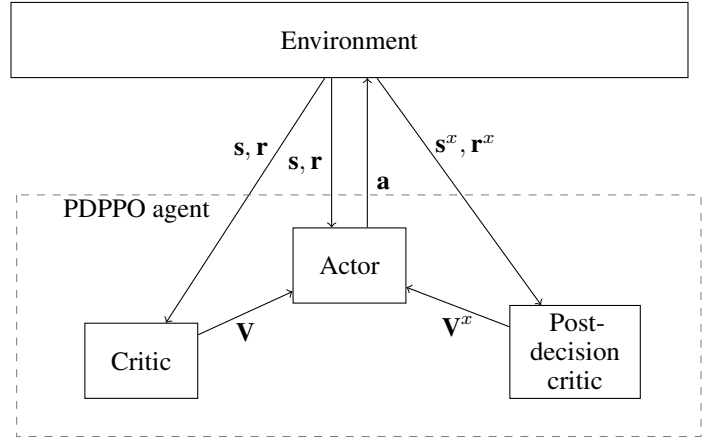

To better show the relation between the two metrics we show in Figure \ref{fig:PDPPO_architecture} the architecture and the interactions between the elements of the PDPPO architecture. The agent interacts with the environment, and each state is evaluated by two critic networks: a Critic and a Post-decision Critic. 
The actor-network is updated by evaluating the value of both the post-decision critic and the critic.
In addition, the agent employs a batch sample consisting of the current state $\textbf{s}$ (here we use the bold notation for a batch of a particular variable), the next state $\textbf{s}^x$, and the reward $\textbf{r}$ to update both the critic and post-decision critic.

Additionally, in the batch generation process, only the critic network is employed to evaluate the post-decision state and state, whereas the post-decision critic network is not used.

\begin{algorithm}
\footnotesize 
\caption{Post-Decision Proximal Policy Optimization (PDPPO)}
\label{algo:PDPPO}
\begin{algorithmic}[1]
\Procedure{PDPPO}{$N,K, \epsilon$}
\LineComment{$N$: Training epochs, $K$: Policy updates/epoch, $\epsilon$: Clipping range}
\LineComment{Output: Trained policy parameters $\theta_\pi$}
\For {$i \in\{1, 2, \ldots, N\}$}
    \State Collect trajectories $D$ using the current policy
    \For {$j \in\{1, 2, \ldots, D\}$}
        \State Compute returns $R$, $R^x$ and advantages $A^{\pi,pre}$, $A^{\pi,x}$ for all states
        \State Compute final advantages $A^\pi = \max(A^{\pi,pre}, A^{\pi,x})$
    \EndFor
    \For {$k \in\{1, 2, \ldots, K\}$}
        \State Sample batch of transitions from $D$
        \State Compute log probabilities, state values, and advantages
        \State Compute importance sampling ratios $\rho_t$
        \State Update policy via actor loss and clipped gradients
        \State Update value function $V^{\pi,x}$ via first critic loss
        \State Update post-decision value function $V^{\pi,pre}$ via second critic loss
    \EndFor
    \State Update old policy $\theta_{\pi,i-1}$ with current policy parameters $\theta_{\pi,i}$
\EndFor
\State \Return{$\theta_{\pi,N}$}
\EndProcedure
\end{algorithmic}
\end{algorithm}

\section{Computational experiments}
\label{sec:computational_experiment}

This section evaluates the performance of PDPPO against other methods. 
We test and compare the PDPPO on the Frozen Lake Environment in Section \ref{subsec:frozen_lake} and the Stochastic Discrete Lot-sizing Environment \ref{subsec:discrete_lot_sizing}. 
Each result is obtained using multiple independent runs to establish the average performance. 
Specifically, this approach provides a robust framework for statistical analysis, including t-tests to confirm the significance of performance differences observed between the models. 
Moreover, in line with the methodological rigor suggested by \cite{Patterson2023}, we conduct enough trials to achieve statistical significance of the difference—30 runs in the Frozen Lake environment and 20 in the lot-sizing environment—ensuring robust comparisons. 

We conducted the experiments on a machine with an AMD Ryzen 5 5600X 6-Core Processor 3.70 GHz, an NVIDIA GeForce RTX 3060 GPU (12GB), and 32GB of RAM. We used PyTorch version 1.8.0 for implementing and training the algorithms \cite{Paszke2019}.

The implementation of all the algorithms, the environments, the hyperparameters, and necessary scripts for running the experiments and plotting the results are available on \url{https://github.com/leokan92/pdppo}. 

The hyperparameters used were uniformly applied across all models to maintain consistency in the comparative analysis. 
This uniformity is essential as it directly compares the algorithms' core performance without variations due to different tuning parameters. 
We selected hyperparameters that demonstrated stable learning curves in previous studies, applying these hyperparameters to the methods studied here. 
This approach ensures that any observed differences in performance can be attributed primarily to the algorithms' inherent strategies rather than external configuration variables. 
While we did not conduct an exhaustive hyperparameter tuning process, we relied on commonly used values established in similar environments, such as those found in Atari games. 
For reference, readers can explore pretrained reinforcement learning baselines with extensively tuned hyperparameters at \url{https://stable-baselines3.readthedocs.io/en/master/guide/rl_zoo.html}. 
The only exception was the factor $\gamma$, for which we conducted a few targeted experiments and identified that a value of $0.90$ yielded the best performance results.

\subsection{Frozen Lake Environment}\label{subsec:frozen_lake}

The Frozen Lake environment, widely used for testing reinforcement learning algorithms, introduces a challenging, sparse reward environment.
We incorporate a stochastic component due to its slippery dynamics. 
This component adds uncertainty to the effects of the agent's actions, making learning more difficult.
This environment is commonly used and is fully reproducible, ensuring reliable experimentation \cite{Gupta2021}.

The agent must navigate the frozen lake to reach a goal while avoiding holes in the ice. 
The environment includes a slippery effect, which adds stochasticity to the agent's movements, making it more challenging to navigate the ice. 
The state space is represented by a $n \times m$ grid, and the action space consists of four possible actions: move up, down, left, or right.
Rewards are $+1$ for reaching the goal and $-1/(n \times m)$ for falling into a hole, a modified dynamic designed to demonstrate the benefits of the proposed method. 
The average reward obtained during the training measures the agent's success rate.
Here, we adopted a $n = m = 10$ grid size, randomly generating a hole in the grid with a probability of $80\%$ of tiles being a hole every new experiment training iteration.

\begin{figure}[htbp]
    \centering
    \includegraphics[width=0.48\textwidth]{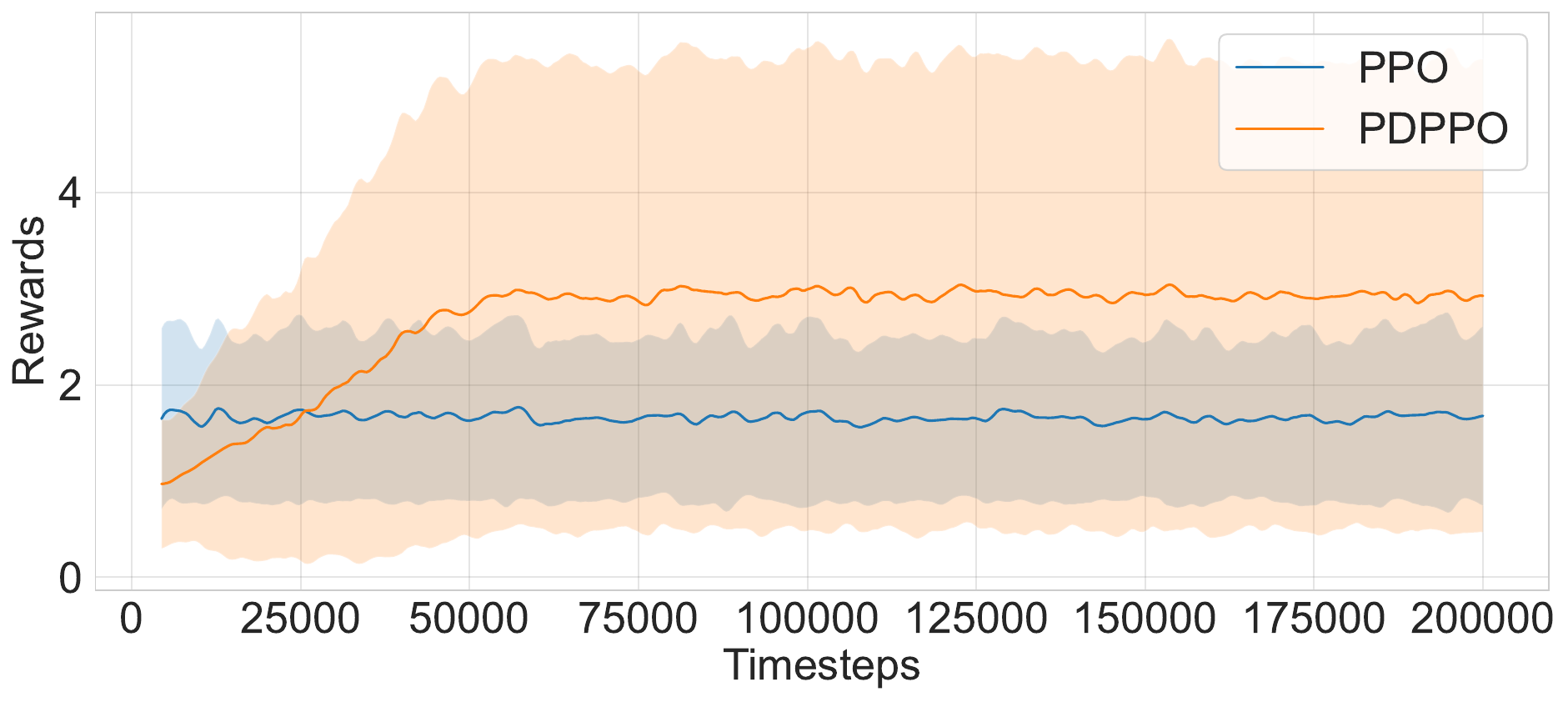}
    \caption{Evolution of rewards in the modified Frozen Lake environment comparing PDPPO and vanilla PPO. The shaded areas around the plot lines represent the 95\% confidence intervals, which are calculated from the distribution of rewards.}
    \label{fig:frozen_lake_results}
\end{figure}

The experiments show that the PDPPO agent outperforms the traditional PPO agent in the Frozen Lake environment. 
We train each method introduced in this paper $30$ times.
As we observe in Figure \ref{fig:frozen_lake_results}, the average value of the reward remains above the average value of the PPO in the vast majority of the runs. 
Notably, the reward evolution curve from PDPPO illustrates a clearer learning dynamic, with rewards progressively increasing as training progresses. 
In contrast, the rewards for the PPO method remain relatively stable, showing minimal improvement across different training steps. 

In Table \ref{tab:frozen_lake_results}, we provide a more detailed breakdown of the performance of these two methods.

\begin{table}[ht]
        \centering
        \caption{The table presents the average and standard deviation of the maximum reward achieved and total cumulative reward for the RL agents PDPPO and PPO, calculated from thirty independent runs. Asterisks denote statistical significance ($p < 0.05$) favoring the model with the higher average reward.}
        \begin{tabular}{l p{2.0cm} p{3.3cm}}
        \toprule
        \textbf{Method} & \textbf{Max Reward (Mean $\pm$ SD)} & \textbf{Total Cumulative Reward (Mean $\pm$ SD)} \\ \midrule
        PDPPO & $5.106 \pm 2.779^{*}$ & $1334.065 \pm 1057.226^{*}$ \\
        PPO & $3.704 \pm 2.083$ & $828.920 \pm 419.150$ \\ \bottomrule
        \end{tabular}
    \label{tab:frozen_lake_results}
\end{table}

It is possible to notice that PDPPO achieves a significantly higher total cumulative reward, averaging $1334.065$ with a standard deviation of $1057.226$, as marked by the asterisks indicating statistical significance. 
These results suggest superior overall performance throughout the run. 
The average maximum reward for PDPPOs was also significantly higher than that of PPOs. 
Moreover, the highest maximum reward for PDPPO is $5.106$, though higher than PPO's, which is $3.704$, comes with a higher standard deviation, reflecting a less consistent performance across different runs.

These findings underscore the effectiveness of the PDPPO approach, which can potentially be attributed to the incorporation of post-decision states and a dual critic network that enhances the method's robustness and reliability despite its varied initializations. 
The statistically significant better results in both maximum reward and total cumulative reward for PDPPO indicate that its adaptation to the PPO framework results in notable performance improvements in challenging environments.

\subsection{Stochastic Discrete Lot-sizing Environment}\label{subsec:discrete_lot_sizing}

As the second environment, we use the Stochastic Discrete Lot-sizing Problem, which addresses the production planning of a number of items in a horizon planning divided into small-time buckets, such as hours or shifts, where parallel machines produce to meet stochastic demand as in \cite{Felizardo2024}. 
The objective is minimizing the setup costs, holding costs, and lost sales penalties, while adhering to constraints such as capacity limits, inventory balancing, and sequence-independent setup times and costs.

\begin{figure}[!ht]
    \centering
    \resizebox{0.5\textwidth}{!}{ 
    \begin{tikzpicture}
        \draw[ultra thick, ->] (-2,0) -- (10,0);
        
        \draw[ultra thick] (-1.5,-0.2) -- (-1.5,0.2);
        \draw[ultra thick, ->] (-1,-1) -- (-1,0);
        \draw[ultra thick, ->] (1,1) -- (1,0);
        \draw[ultra thick, ->] (3,-1) -- (3,0);
        \draw[ultra thick, ->] (5,1) -- (5,0);
        \draw[ultra thick, ->] (7,-1) -- (7,0);
        \draw[ultra thick] (8,-0.2) -- (8,0.2);
        \draw[ultra thick, dashed, ->] (9,1) -- (9,0);
        
        \draw[ultra thick] (-1, -1) node[below=5pt,thick] {$d_t$} node[below=4pt] {};
        \draw[ultra thick] (1,1) node[above=5pt,thick] {change setup} node[above=4pt] {};
        \draw[ultra thick] (3,-1) node[below=5pt,thick] {production} node[above=4pt] {};
        \draw[ultra thick] (5,1) node[above=5pt,thick] {satisfy demand $d_{t}$} node[above=4pt] {};
        \draw[ultra thick] (7,-1) node[below=5pt,thick] {pay lost sales and holding costs} node[above=4pt] {};
        \draw[ultra thick] (9,1) node[above=5pt,thick] {$d_{t + 1}$} node[above=4pt] {};
        \draw[ultra thick] (-1.5,0) node[above=5pt,thick] {$t$} node[above=4pt] {};
        \draw[ultra thick] (8,0) node[above=5pt,thick] {$t + 1$} node[above=4pt] {};
    \end{tikzpicture}
    }
    \caption{Flow of operations}\label{fig:informationFlow}
\end{figure}
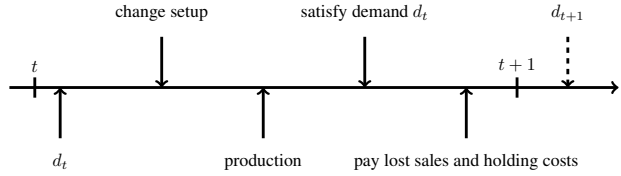

If a change in the machine configuration is required, a setup cost is incurred, and items are produced according to each machine capacity. 
After production, the available items are used to satisfy demand $d_t$, and both lost sales and holding costs are computed. 
In particular, for each item, we have holding costs and lost sales costs. The flow of events is represented in Figure \ref{fig:informationFlow}.

We model the DLSP by means of a \textit{Markov Decision Process}. 
Thus, we define \textit{states}, \textit{actions}, \textit{immediate contribution} and \textit{state transition}. 
We refer the reader to \cite{Felizardo2024} for a more detailed description of the environment.

The \textit{state} is described by:
\begin{itemize}
    \item $\mathbf{I}_t = [I_{1,t}, \dots, I_{L,t}]$: the array of inventory levels, where $I_{i,t}$ is the inventory of item $i$ at time $t$.
    \item $\mathbf{M}_t = [M_{1,t}, \dots, M_{Z,t}]$: the array of machine configurations, where $M_{j,t}\ \in\ \mathcal{I}(m)$ is the configuration of machine $j$ at time $t$.
\end{itemize} 

The \textit{actions} are described by $\mathbf{A}_{t} = [a_{i,j,t}]_{M\in \mathcal{M}, I \in \mathcal{I}}$, binary variables equal to 1 if machine $j$ is producing item $i$ at time $t$. 
Moreover, for modeling purposes, the following variables are defined: $\delta_{i,j,t}$ is a binary variable indicating if machine $j$ is set up for item $i$ between $t-1$ and $t$; $d_{i,t}$ represents the demand of item $i$ in period $t$; $f_{i}$ is the fixed setup cost for item $i$; $h_{i}$ is the unit holding cost for item $i$; $I_{i,t}$ denotes the inventory level of item $i$ at the beginning of period $t$; $p_{i,j}$ is the production capacity of machine $m$ for item $i$; $c_{i,j}$ is the cost of changing the setup of machine $m$ for item $i$; $l_{i}$ is the backorder cost for item $i$; and $[y]^{+} = \max\{y, 0\}$ represents the positive part of $y$. The \textit{immediate contribution} depends on setup costs, ending inventory, and potential lost sales.

The \textit{immediate contribution} depends on the setup cost, the inventory at the end of the time period, and the possible lost sales:

\begin{equation}
\label{eq:cost}
\begin{aligned}[t]
    R_{t}\left(\mathbf{X}_{t}, \mathbf{I}_{t}, \mathbf{M}_{t}\right) = 
    \sum_{j=1}^{Z} \sum_{i=1}^{N} f_{i} \delta_{i,m,t} + \\
    \sum_{i=1}^{L} h_{i} \Bigg[I_{i,t} + 
    \sum_{j=1}^{Z} \big(p_{i,j} a_{i,j,t} - c_{i,j} \delta_{i,j,t}\big) - d_{i,t} \Bigg]^{+} + \\
    \sum_{i=1}^{L} l_{i} \Bigg[d_{i,t} - I_{i,t} + 
    \sum_{j=1}^{Z} \big(p_{i,j} a_{i,j,t} - c_{i,j} \delta_{i,j,t}\big) \Bigg]^{+}.
\end{aligned}
\end{equation}

Finally, the \textit{state transition} equation for inventory is:

\begin{equation}
    \label{eq:state_tran_inven}
    \ I_{i, t+1}=\left[I_{i,t}+\sum_{j=1}^{Z}\left(p_{i,j} a_{i,j,t}-c_{i,j} \delta_{i,j,t}\right)-d_{i, t}\right]^{+}.
\end{equation}

The environment poses several challenges for reinforcement learning (RL) methods, summarized as follows:

\begin{itemize}
    \item \textbf{Large state and action space}: The state space includes inventory levels, and the action space involves binary variables for machine-item production. These grow exponentially with the number of items, machines, and time periods, making traditional tabular RL methods (e.g., Q-learning \cite{Watkins1989}, SARSA \cite{Rummery1994}) infeasible.
    \item \textbf{Stochastic demand}: Item demand is uncertain, introducing stochasticity in state transitions and rewards. RL algorithms must adapt to varying demand patterns to learn effective policies.
    \item \textbf{Long-term dependencies}: Current decisions can impact future states and rewards, such as production losses from machine setups or sales losses from unmet demand. RL must account for these dependencies to optimize policies.
    \item \textbf{Sparse, delayed, and mixed rewards}: Rewards are often sparse, delayed, and a mix of deterministic (e.g., machine setups) and stochastic (e.g., stockouts). This caracteristic makes credit assignment and learning effective policies more complex, especially when decisions have long-term and interwoven impacts.
\end{itemize}

We also define a variation of PDPPO, called PDPPO1C, in which a single critic network is employed.
In this environment, we can perform an ablation study better regarding including the extra critic network focused on the post-decision states.
In the lot-sizing environment, we have a higher-dimensionality state and action space that adds complexity to the problem and helps us evaluate the impact of the additional critic network.
This critic is fed just with the post-decision states. 
Moreover, comparing  PDPPO and PDPPO1C allows us to measure the value of the dual policy concerning having just one.

\begin{figure*}[ht]
  \centering
  \begin{subfigure}{0.3\linewidth}
    \includegraphics[width=\linewidth]{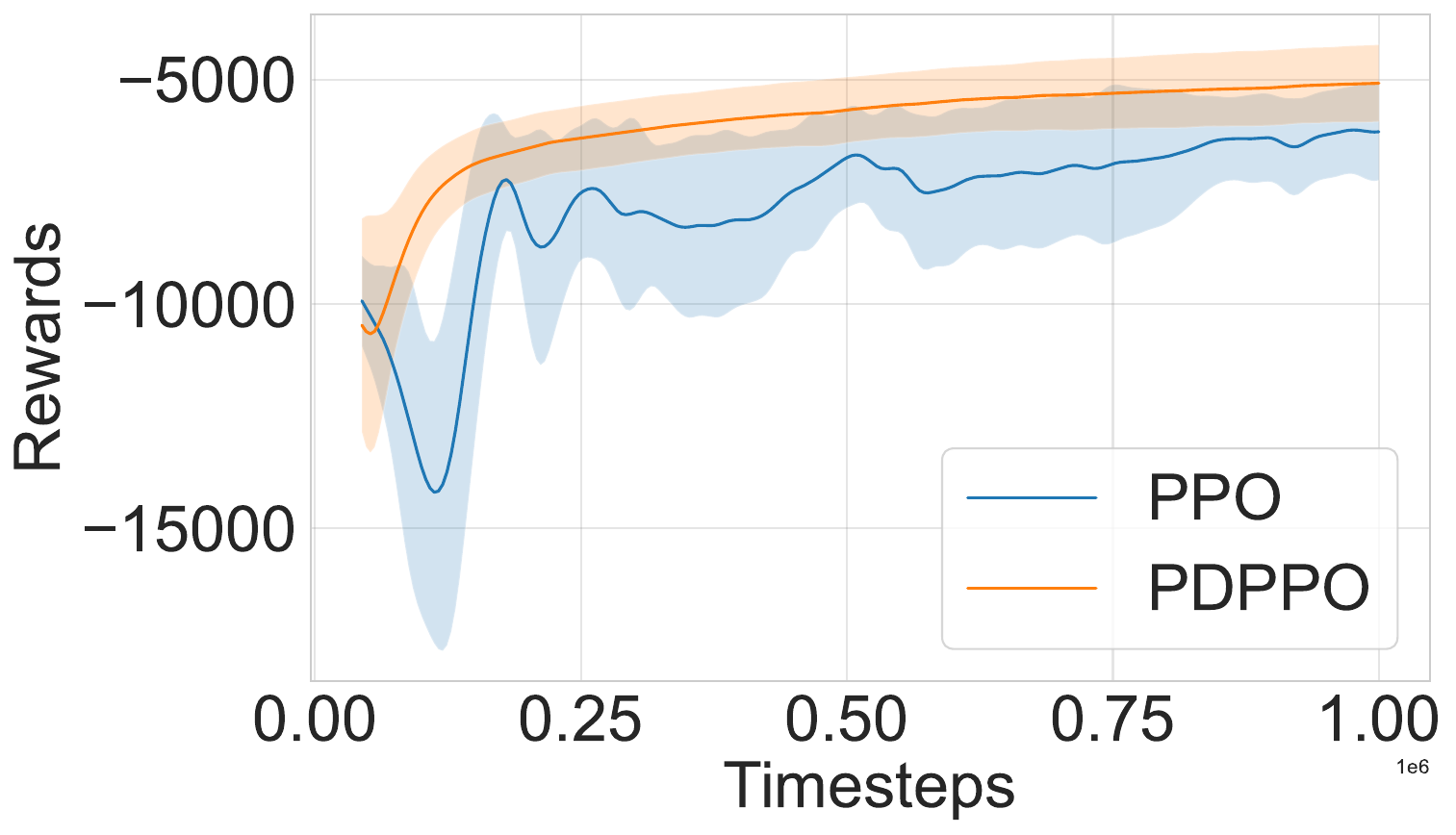}
    \caption{Environment configuration: 20 items, 10 machines, 100 maximum inventory capacity}
    \label{fig:rewards1}
  \end{subfigure}
  \hfill
  \begin{subfigure}{0.3\linewidth}
    \includegraphics[width=\linewidth]{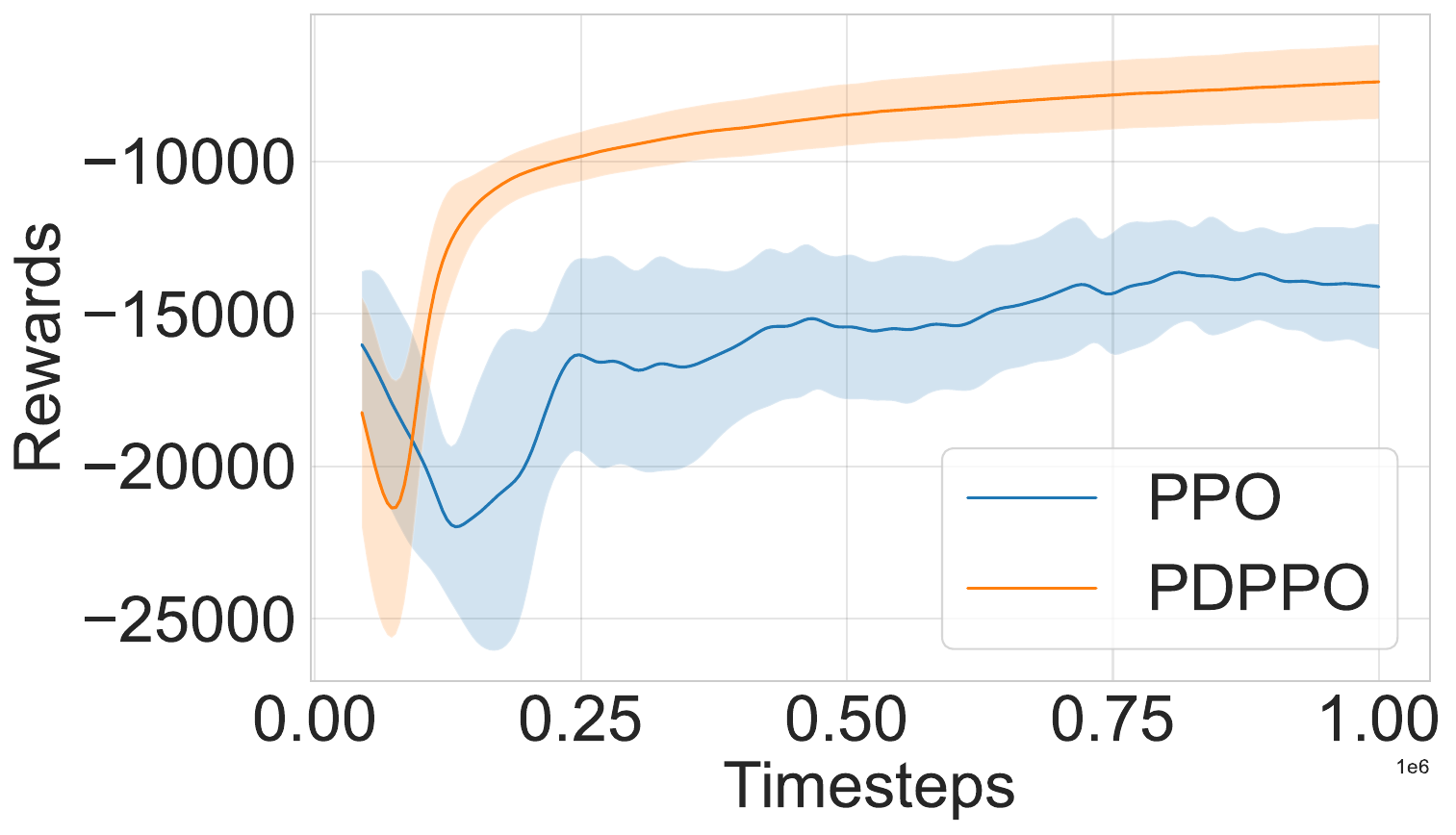}
    \caption{Environment configuration: 25 items, 10 machines, 100 maximum inventory capacity}
    \label{fig:rewards2}
  \end{subfigure}
  \hfill
  \begin{subfigure}{0.3\linewidth}
    \includegraphics[width=\linewidth]{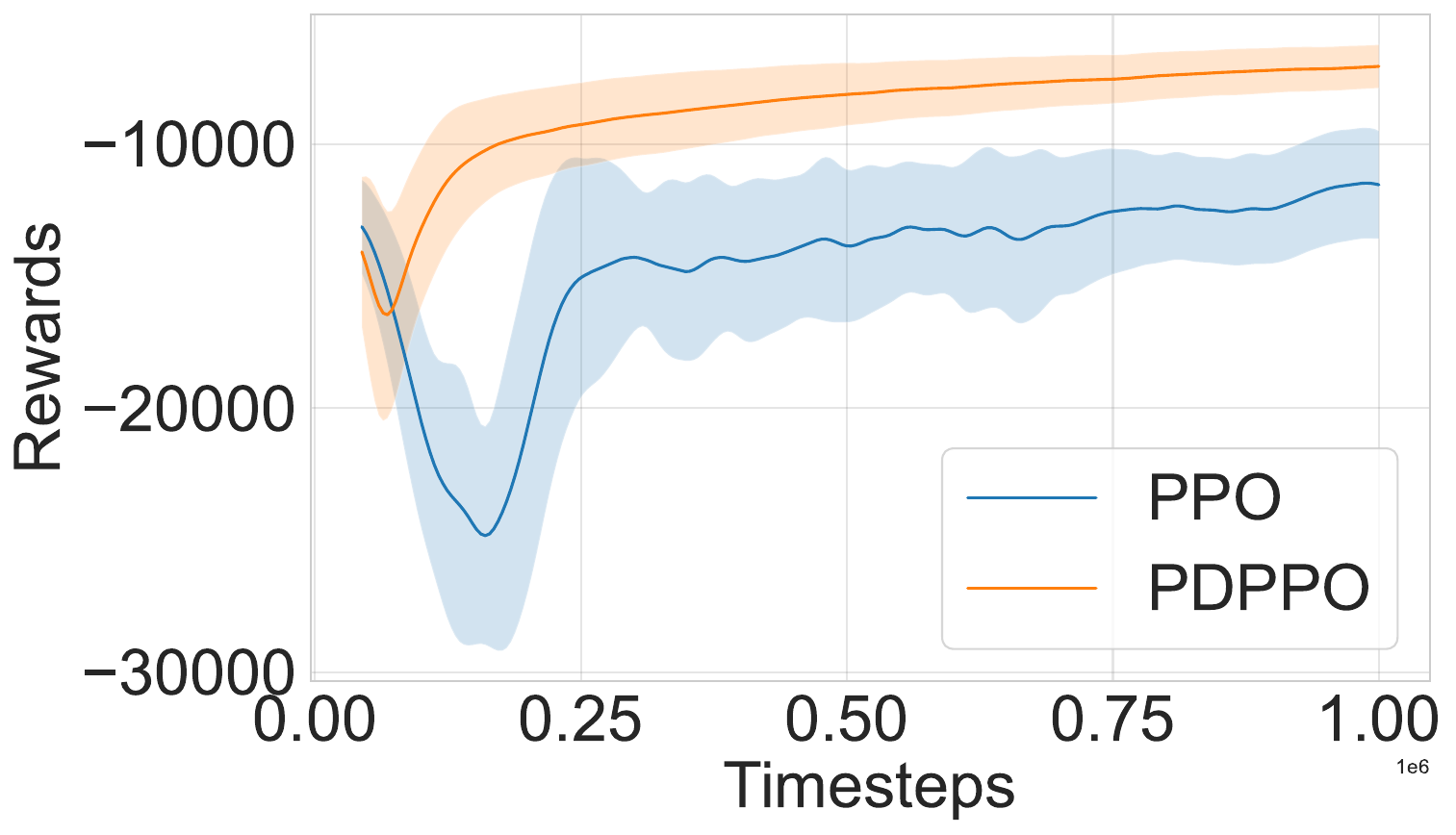}
    \caption{Environment configuration: 25 items, 10 machines, 100 maximum inventory capacity}
    \label{fig:rewards3}
  \end{subfigure}
  \caption{Evolution of rewards over time for PDPPO with dual critic, and standard PPO for three different environment configurations. The shaded areas around the plot lines represent the 95\% confidence intervals, which are calculated from the distribution of rewards.}
  \label{fig:rewards_PDPPO}
\end{figure*}

In this environment, we trained PDPPO, PDPPO1C, and PPO, and we report the evolution of the performance metric values in Figure \ref{fig:rewards_PDPPO}. 
We tested different environment configurations with varying dimensionality, ranging from medium to higher dimensions, as shown in Figures \ref{fig:rewards1}, \ref{fig:rewards2}, and \ref{fig:rewards3}. 
These configurations include 20 items with 10 machines, 25 items with 10 machines, and 25 items with 15 machines, respectively. 
The experiments presented in this paper complement the scenarios analyzed in \cite{Felizardo2024}, which also explores similar stochastic lot-sizing setups focusing on medium and higher dimensionalities. 
As shown in Figure \ref{fig:rewards1}, PDPPO outperforms PPO regarding faster learning and achieving higher reward levels, having PPO a higher standard deviation. This trend is also observed, but more accentuated, in Figure \ref{fig:rewards2} and \ref{fig:rewards3}, where the problem's dimensionality is higher.

\begin{table*}[ht]
\centering
\caption{Consolidated performance metrics across environments. Asterisks denote the level of significance in statistical tests comparing PDPPO to the second best method. A double asterisk (**) indicates $p < 0.01$, denoting significant differences at the 1\% level.}
\label{tab:res_SDLSP}
\begin{tabular}{@{}llll@{}}
\toprule
\textbf{Environment}                  & \textbf{Method}                 & \textbf{Max Reward (Mean $\pm$ SD)}                  & \textbf{Cumulative Reward (Mean $\pm$ SD)}  \\ \midrule
20 Items and 10 Machines          & \textbf{PDPPO}                           & $\mathbf{-4.995 \times 10^3 \pm 8.116 \times 10^2}$          & $\mathbf{-1.510 \times 10^6 \pm 1.5455 \times 10^5}^{**}$ \\
                                   & PPO                             & $-5.112 \times 10^3 \pm 5.860 \times 10^2$          & $-1.946 \times 10^6 \pm 2.5386 \times 10^5$ \\
                                   & PDPPO1C                       & $-5.958 \times 10^3 \pm 5.772 \times 10^2$          & $-2.101 \times 10^6 \pm 2.2989 \times 10^5$ \\
                                   \midrule
25 Items and 10 Machines          & \textbf{PDPPO}                           & $\mathbf{-7.291 \times 10^3 \pm 1.1851 \times 10^3}^{**}$       & $\mathbf{-2.394 \times 10^6 \pm 1.8624 \times 10^5}^{**}$ \\
                                   & PPO                             & $-10.779 \times 10^3 \pm 1.4174 \times 10^3$        & $-3.973 \times 10^6 \pm 3.6527 \times 10^5$ \\
                                   & PDPPO1C                       & $-9.195 \times 10^3 \pm 1.2007 \times 10^3$         & $-2.861 \times 10^6 \pm 4.0851 \times 10^5$ \\
                                   \midrule
25 Items and 15 Machines          & \textbf{PDPPO}                           & $\mathbf{-6.938 \times 10^3 \pm 7.938 \times 10^2}^{**}$        & $\mathbf{-2.197 \times 10^6 \pm 3.0365 \times 10^5}^{**}$ \\
                                   & PPO                             & $-9.142 \times 10^3 \pm 1.6590 \times 10^3$         & $-3.650 \times 10^6 \pm 5.6793 \times 10^5$ \\
                                   & PDPPO1C                       & $-8.501 \times 10^3 \pm 8.588 \times 10^2$          & $-3.001 \times 10^6 \pm 2.1630 \times 10^5$ \\ \bottomrule
\end{tabular}
\end{table*}

In Table \ref{tab:res_SDLSP}, we compare the performance of PDPPO with PDPPO1C and PPO across different environments and configurations. 
All the results are averaged and computed on $20$ experiments per configuration, providing a robust dataset sufficient for statistically significant differences.

For the \textit{20 Items and 10 Machines} environment, PDPPO outperforms PPO; the difference is not statistically significant for the Max Reward, indicated by a p-value greater than 0.05. However, when analyzing the Cumulative Reward, we achieved statistical significance. Despite a less distinct edge in maximum rewards, PDPPO continues to sustain a lead in cumulative rewards. 

In the larger \textit{25 Items and 10 Machines} environment, the difference between the maximum rewards achieved by PDPPO and PPO is statistically significant (indicated by **), highlighting a pronounced advantage for PDPPO in handling increased complexity in both metrics, Max Reward and Cumulative Reward.

Finally, in the \textit{25 Items and 15 Machines} environment, PDPPO significantly outperforms PPO in both maximum and cumulative rewards. This significant improvement underscores PDPPO's effective utilization of dual critics, especially in more complex configurations. 
The performance data support PDPPO and give evidence of the marginal increase of the PDPPO Max Rewards and Cumulative Rewards as the environment dimensionality grows. 
This evidence means that PDPPO can maintain effectiveness with increasingly larger problem dimensions.

These results suggest that PDPPO with dual-critic is more efficient than PPO for solving the stochastic lot-sizing problem. It is worth noting that PDPPO1C, which utilizes a single critic, shows varied performance across different environments. 
In the \textit{20 Items and 10 Machines} environment, PDPPO1C under-performs compared to both PDPPO and PPO, with the lowest mean maximum reward and cumulative reward. 
These results suggest that while the dual critic design of PDPPO enhances the algorithm's efficiency and ability to interpret complex environments, the single critic approach in PDPPO1C may not capture enough complexity to handle the task effectively in smaller or less complex settings.
However, the results in the environments \textit{20 Items and 15 Machines} and \textit{25 Items and 15 Machines} provide a different perspective. 
Although PDPPO1C still does not match the performance of PDPPO, it significantly improves its performance in smaller environments and even outperforms the standard PPO in larger settings. 
This is reasonable since, by using post-decision state variables, PDPPO1C can better estimate the state value and thus be more efficient in learning. 
From the observed results, we might conclude that incorporating the post-decision state information may provide a more accurate representation of the current state of the environment, which could improve the agent's decision-making process. 

\section{Conclusions}\label{sec:conclusion}
In this study, we introduced a variant of the Proximal Policy Optimization (PPO) algorithm—coined Post-Decision Proximal Policy Optimization (PDPPO)—with a dual critic specifically tailored for problems with a deterministic and a stochastic component in the state transition function. 
This variant integrates post-decision states and dual critics, effectively giving the agents more information about the environment dynamics and enhancing the learning process. 
Our empirical analysis demonstrates that PDPPO with dual critic performs better than standard PPO and PDPPO with a single critic. 
Importantly, our method achieves faster learning and higher average rewards and demonstrates lower reward variability. This stability, independent of neural network initialization and the problem's initial state, underscores the reliability of the PDPPO algorithm.

The versatility of the PDPPO with dual critic architecture, especially in complex, high-dimensional stochastic environments, could be promising for applications in several fields, such as robotics, finance, telecommunication, and gaming.
Incorporating post-decision states alongside dual critics has shown some evidence to be a powerful tool. Future research could explore the adaptability of this approach across these domains, benchmarking against other cutting-edge algorithms. 

In conclusion, our study underscores the importance of exploring novel techniques and architectures in reinforcement learning. This approach is crucial in tackling the challenges posed by complex and dynamic environments, and we hope that our findings will inspire further research in this direction.


\section*{Acknowledgements}
The authors are also grateful for the financial support provided by CNPq (403735/2021-1; 309385/2021-0) and FAPESP (2013/07375-0; 2022/05803-3).


\bibliographystyle{plain}

\bibliography{Bibliography_RS}

\end{document}